# Heterogeneous virus classification using a functional deep learning model based on transmission electron microscopy images

**Niloy Sikder[1,2], Md. Al-Masrur Khan[3], Anupam Kumar Bairagi[4], Mehedi Masud[5], Jun Jiat Tiang[6*], Abdullah-Al Nahid[7*]**

[1] Radboud University Medical Center, Donders Institute for Brain, Cognition and Behaviour, Nijmegen, The Netherlands.
[2] Faculty of Technology and Bionics, Rhine-Waal University of Applied Sciences, Kleve, Germany.
[3] Md. Al-Masrur Khan PhD student, Department of Intelligent Systems Engineering, Indiana University Bloomington, IN, USA.
[4] Computer Science and Engineering Discipline, Khulna University, Khulna 9208, Bangladesh.
[5] Department of Computer Science, College of Computers and Information Technology, Taif University, P.O. Box 11099, Taif 21944, Saudi Arabia.
[6] Centre For Wireless Technology (CWT), Faculty of Engineering, Multimedia University, Cyberjaya-63100, Malaysia.
[7] Electronics and Communication Engineering Discipline, Khulna University, Khulna 9208, Bangladesh.

[*]Corresponding Authors: Abdullah-Al Nahid (nahid.ece.ku@gmail.com) and Jun Jiat Tiang (jjtiang@mmu.edu.my)

## Abstract

Viruses are submicroscopic agents that can infect all kinds of lifeforms and use their hosts' living cells to replicate themselves. Despite having some of the simplest genetic structures among all living beings, viruses are highly adaptable, resilient, and given the right conditions, are capable of causing unforeseen complications in their hosts' bodies. Due to their multiple transmission pathways, high contagion rate, and lethality, viruses are the biggest biological threat faced by animal and plant species. It is often challenging to promptly detect the presence of a virus in a possible host's body and accurately determine its type using manual examination techniques; however, it can be done using computer-based automatic diagnosis methods. Most notably, the analysis of Transmission Electron Microscopy (TEM) images has been proven to be quite successful in instant virus identification. Using TEM images collected from a recently published dataset, this article proposes a deep learning-based classification model to identify the type of virus within those images correctly. The methodology of this study includes two coherent image processing techniques to reduce the noise present in the raw microscopy images. Experimental results show that it can differentiate among the 14 types of viruses present in the dataset with a maximum of 97.44% classification accuracy and $F_1$-score, which asserts the effectiveness and reliability of the proposed method. Implementing this scheme will impart a fast and dependable way of virus identification subsidiary to the thorough diagnostic procedures.

## 1 Introduction

Transmission Electron Microscopy (TEM) refers to an imaging technique in which electron microscopes are used to transmit a beam of electrons through specimens to produce detailed and high-resolution microscopic images. Max Knoll and Ernst Ruska were the first to construct this type of microscopes in the early 1930s [1]. TEM quickly became a prominent and widely-used imaging technique because of its nanometer-scale (often even sub-nanometer-scale) resolution, allowing direct visualization of molecules, cells, tissue sections, and tiny microorganisms and pathogens, such as viruses and bacteria [2]. In the history of virology, TEM has been influential in discovering countless new viruses and establishing their taxonomies [3]. It is significantly better than other microscopy imaging techniques, such as Bright-field Microscopy (the theoretical resolution limit is 0.2μm), Fluorescence Microscopy (can capture only single-color images), and High-Content Screening [4]. In virology, TEM serves as a routine virus diagnostic technique and a critical tool for infectious agent detection in the occurrence of a new (and unusual) disease outbreak or possible bioterrorism attacks [3]. It is often called a *catch-all* method, as it can be used to detect multiple types of viruses present in a single sample, even without the help of any *a priori* information [5]. However, the biggest hurdle of using TEM imaging for virus diagnosis is that it is over-dependent on skilled and experienced virologists, pathologists, or other trained professionals, which makes it costly, time-consuming, and quite challenging to rely on during outbreaks (when mass-diagnosis is required). It is also not possible to identify a virus species beyond the genus level by analyzing its morphological appearance (in a TEM image); however, experts can narrow down the possibilities and run other molecular tests for more specific genera or strains [5]. During the 1990s, more sophisticated techniques like Enzym-Linked Immunosorbent Assay (ELISA) and Polymerase Chain Reaction (PCR) were developed to improve the diagnosis of viral infections considerably [6]. Nevertheless, despite all its drawbacks, TEM image analysis remains one of the standard diagnostic tools in our relentless fight against existing and new pathogens. In a 2019 article, Roingeard et al. [2] discussed the history, contributions, and current roles of TEM imaging in virology at length. Apart from virology, the technology has a broad range of applications in materials science, nanotechnology, and other branches of biological, physical, chemical sciences.

In recent years, another set of applications of TEM virus images has emerged due to the surge in usage of Machine Learning (ML) techniques in Biomedical science, especially in disease detection and diagnosis. The field of Computer-Aided Diagnosis (CAD) has grown rapidly in the last two decades, and its advancements are expected to be implemented in mainstream medical tests within a few years. ML has a number of powerful and robust algorithms to teach machines how to make decisions like a human. Coupled with the existing Digital Image Processing (DIP) algorithms, these methods can identify various species in TEM images quite effectively. Especially the Convolutional Neural Network (CNN)-based deep learning models have proven to be very effective in TEM image classification. In the last few years, several groups of researchers have proposed and validated such automatic diagnosis methods. However, their performance still has some shortcomings, making them less useful for practical use. Apart from virus classification, TEM virus images have been used for virus particle detection [6], segmentation of virus particle candidates [5], bacteriophage morphological characterization [7], prediction of cell migration [8], and identification of virus structures [9]. Deep learning models are also being used on other medical image data for automatic diagnostic purposes, including chest X-ray images to detect pneumonia [10] and COVID-19 cases [11], histopathology images to detect Invasive Ductal Carcinoma (IDC) [12] and Breast Cancer cases [13], blood smear images to diagnose malaria parasites [14], and dermoscopy images for skin lesion segmentation [15] and to identify various skin diseases [16].

Upon exploring the existing literature, a recent study by Zhang et al. [17] suggests that, in microbiology, deep learning-based methods are yet to reach their full potential. They believe, if the professionals of the field embrace these methods, it will open up another perspective and lead to new applications and discoveries. In this study, we aim to develop a functional deep learning model for fast and automatic virus classification by collecting features from a set of TEM images (of multiple virus species). We used two popular DIP algorithms, namely local standard deviation filtering and two-dimensional Discrete Cosine Transform (2D DCT), for noise removal and sample preparation. Finally, we used a customized CNN architecture to extract convolutional features from the samples and classify them. The proposed method and the results obtained by it have been described and discussed elaborately with necessary graphs, tables, figures, and other illustrations.

The rest of the paper is organized as follows. Section 2 presents a chronological discussion on some of the major works in this area of research. Section 3 outlines our virus classification methodology and elaborates on each step of the workflow. Section 4 presents the acquired results, provides discussions on them, and compares the method's performance with other studies. Finally, Section 5 summarizes the takeaways of the article with some concluding remarks and future research plans.

## 2 Literature Review

In this section, we take a look at the existing ML methods for virus identification based on TEM images. One of the biggest challenges of working in this research area is the lack of (multiple) labeled datasets. It is often challenging to collect a large number of diverse samples (of numerous viruses) from various sources, go through the (standard) procedure of preparing them for scans, have access to electron microscopes (to capture the TEM images), and assemble a team of experts to grade them cautiously — all of which are necessary steps to accumulate a good dataset. In 2011, Kylberg et al. put together a TEM dataset [18], which is still considered a benchmark for building and testing ML-based virus diagnosis methods. The dataset is named *Virus Texture Dataset*, and it contains 1500 TEM images (greyscale, 41 $\times$ 41 pixels) evenly distributed to 15 virus classes [19]. Most of the works (in this field) within the last decade use the microscopic virus images of this dataset. Bellow, we present concise descriptions of some of the notable virus classification methods published in recent years.

In 2014, Sintorn et al. [20] presented a virus recognition scheme by analyzing TEM images based on five types of local texture features. Their study used two feature selection methods (forward selection and importance selection) and the Random Forest (RF) classifier for virus identification. They used the Virus Texture Dataset (described above) along with non-virus samples to train and validate their model, which achieved a peak classification accuracy of 89%. Nanni et al. [21] used an ensemble of texture descriptors to analyze virus textures in TEM virus images. They described several new multi-quinary (MQ) codings for feature extraction and a feature selection technique to reduce the data dimension. Their approach involves two Support Vector Machines (SVMs) for classification, in which they achieved a mean accuracy of 85.70% on the samples of the Virus Texture Dataset. In the same year, Faraki and Harandi [22] described a texture classification method based on the Bag of Riemannian-Words (BoRW) approach. They used the Virus Texture Dataset for model evaluation and classified the samples of the dataset with a 67.50% mean accuracy with a visual descriptor called Local Binary Patterns (LBP).

In 2016, Wen et al. [23] published a virus image classification method using Multi-scale Principal Component Analysis (PCA) and Multi-scale Completed LBP. After filtering the raw images, they used multi-scale PCA to extract features and a completed LBP (CLBP) descriptor to depict the features collected from all filtered images. They applied their feature extraction method on the Virus Texture Dataset and classified its samples with SVM, which yielded an 86.20% peak accuracy. A 2017 article by Shakri et al. [24] works with entropy features extracted from TEM images for sample classification. They also used a median filter at the pre-processing stage. They used only five classes' data from the Virus Texture Dataset and classified them using a Feed-Forward Neural Network (FFNN). The peak accuracy they recorded from their experiments was 88%. In 2018, Wen et al. [25] described a virus categorization method that incorporates multiple techniques for feature extraction, including PCA, Gaussian Processing, and Histograms. They combined all the collected features and performed the final classification using SVM. Their CLBP Histogram Fourier (CLBP-HF) method achieved a peak accuracy of 88% when SVM was used as the final classifier.

In 2019, Matuszewski and Sintorn [26] used a U-Net-based architecture for pixel-wise virus image classification. The study put emphasis on making the model lighter by reducing the number of trainable weights. The authors used 1077 images (of 15 classes) from the Virus Texture Dataset. They achieved an 82.2% classification accuracy, which is close to what the original U-Net provides, with four times less trainable weights. Another 2019 article by Kumar and Maji [27] uses SVM for multi-class virus classification. Their method considered the relevant modality for each pair of classes of the Virus Texture Dataset and achieved a mean classification accuracy of 82.07% after 10-fold cross-validation.

In 2020, Backes and de Mesquita Sa Junior [28] described a feature fusion approach for TEM image classification. They used numerous texture analysis methods, including Gabor wavelets, LBP, Gray Level Dependence Matrix (GLDM), Fourier descriptors, and Discrete Cosine Transform (DCT). They acquired a mean accuracy of 87.27% using SVM. Another 2020 study by de Geus et al. [29] proposed a transfer learning-based virus species identification approach from a pre-trained deep learning model. While changing the number of epochs, they employed several learning models at the experimental stage, including ResNet, InceptionV3, and SqueezeNet. However, they achieved peak classification performance from DenseNet after training it for 50 epochs. Nanni et al. [30] extracted handcrafted features from the Virus Texture Dataset's TEM images and used the DenseNet201 architecture, pre-trained on ImageNet, for class identification. They used numerous feature extraction techniques to draw out the most significant features and, by combining them, acquired a peak accuracy of 89.47%.

In early 2021, Xiao et al. [1] provided a Residual Mixed Attention Network (RMAN), which focused on channel attention and was trained in an end-to-end manner. The resultant method separates 12 virus types with a top-1 error rate of 4.285%. Majtner et al. [31] used Xception and NASNet, numerous popular deep learning architectures for TEM virus classification on the KTH-TIPS2-b and Virus Texture datasets. They employed a texture-based image transformation technique using Contrast-Limited Adaptive Histogram Equalization (CLAHE) method to improve the classification performance of their method. Jena et al. [32] attempted to separate five types of viruses by analyzing their TEM images. They experimented with five different classifiers — Logistic Regression (LR), a neural network, k-Nearest Neighbors (kNN), and Naive Bayes (NB) — and various numbers of folds (NoF) at the classification stage. They achieved their best performance while using the NB classifier with the NoF value set to 20. Rey et al. [33] reported a study using TEM images for HIV-1 research. They designed a two-stage Region-based Convolutional Neural Network (RCNN) to classify and segment HIV-1 virions at various stages of maturation and morphogenesis.

The discussion above signifies the number of available methods for virus classification from TEM images and their diversity, especially in terms of feature extraction and classification techniques. The mentioned performance underlines that, despite the effort of so many studies, the maximum classification accuracy is still below 90% (in the case of the Virus Texture Dataset). This emphasizes the necessity for a more accurate, robust, and reliable method for virus image classification — a challenge we attempted to meet in this study.

# 3 Methodology

This study follows a relatively straightforward workflow. First, we pre-processed the raw TEM images using two common image processing methods — one for filtering and (separately) another for domain transformation. Then these two sets of output images were fed into two different convolutional models for local feature extraction. And finally, a multi-layer perceptron was used to carry out the virus classification. The entire process has been illustrated in Fig. 1. Detailed descriptions of each of these steps have been provided in the following subsections.

## 3.1 Description of the Dataset

The dataset used in this study is called the *TEM virus dataset*, and it is hosted in Mendeley Data [34]. Published in 2021, it contains microscopic images of numerous viruses captured using two electron microscopes — an LEO with a Morada camera and a Tecnai 10 with a MegaView III camera. Before capturing the images, all samples were treated with ten percent phosphate-buffered saline (PBS), placed on carbon-coated TEM grids, and stained with two percent PhosphoTungstic Acid (PTA). The researchers faced many challenges while assembling this dataset, including limited annotation, diffuse virus boundaries, noise, imperfect focus, different magnifications, and debris appearance. Initially, the dataset was imbalanced in terms of class sample sizes, and the raw microscopic images had various resolutions. However, the researchers solved these problems by cropping the raw images to obtain multiple patches (of 256 × 256 pixels each) following some class-wise varying rules and augmenting the acquired patches to even the sample sizes. The dataset's images are grayscale and provided in Tag Image File (.tif) format. In this study, microscopic images of 14 different viruses were used. The authors of the dataset performed a similar classification in their 2021 publication [35]. They used the DenseNet201 architecture, which was pre-trained on ImageNet, to classify the TEM images. Another 2022 publication by Ali et al. [36] described a customized CNN model to classify the virus classses of the dataset. Table 1 describes various aspects of these viruses and their effects on humans and other animals. It also mentions their sample size (used in this study) along with the assigned class IDs. A sample from each virus class has been provided in Figure 2 for visual comparison.

## 3.2 Image preparation for the First Convolutional Model

As mentioned earlier, we processed the raw images separately with two image processing techniques.
However, first of all, we resized the original images to 128 × 128 pixels — it helped to significantly reduce the size of the data along with the time and space complexities of the deep learning model. For the First Convolutional Model, we filtered the TEM images using the local standard deviation filter. It is a type of spatial filter and operates based on the idea of "moving windows" [37]. A moving window represents a small region of the entire image, which allows recording local properties of that region. The window is moved all across the image to extract those properties from the entire image. If we consider $\boldsymbol{A}$ as a sample grayscale image $x_i$ as the intensity of the $i$th pixel of the active window $\omega$, and $p$ as the number of total pixels in the window, then the mean intensity of the window can be calculated using [38].

$$mean(\boldsymbol{A}_\omega) = \frac{1}{p}\sum_{i=1}^{p} x_i^{\boldsymbol{A}_\omega} \quad (1)$$

The standard deviation expresses the deviation of the intensity values of $\omega$ from the mean value, which can be calculated using:

$$SD(\boldsymbol{A}_\omega) = \sqrt{\frac{1}{p}\sum_{i=1}^{p}\left(x_i^{\boldsymbol{A}_\omega} - mean(\boldsymbol{A}_\omega)\right)^2} \quad (2)$$

If this process is repeated for all the windows, it will result in a matrix containing the standard deviations of the intensities from their local mean values. In this study, we moved a 3 × 3 (pixel) window on the TEM images and used the resultant matrix as the input of our First Convolutional Model. However, using a 3 × 3 window on an image in the aforementioned procedure reduces the resolution of the final matrix by two pixels in each dimension from the original image. To compensate for this loss, first, we padded the images with one extra pixel on each side and then performed the standard deviation filtering. For clarification, we presented a small portion of a TEM virus image (4 × 4 pixels) in Fig. 3(a). Then we used symmetric padding on each side of the image to scale it up (to 6 × 6 pixels), as shown in Fig. 3(b). The padded pixels have been highlighted with a gray background. It is worth mentioning that other padding methods can also be used. Fig. 3(c) presents the mean values acquired by driving a 3 × 3 pixel window on the padded image and using equation (1), which results in a matrix equal to the size of the image shown in Fig. 3(a). And finally, the local standard deviation values were calculated from the previous two matrices using equation (2), which is shown in Fig. 3(d).

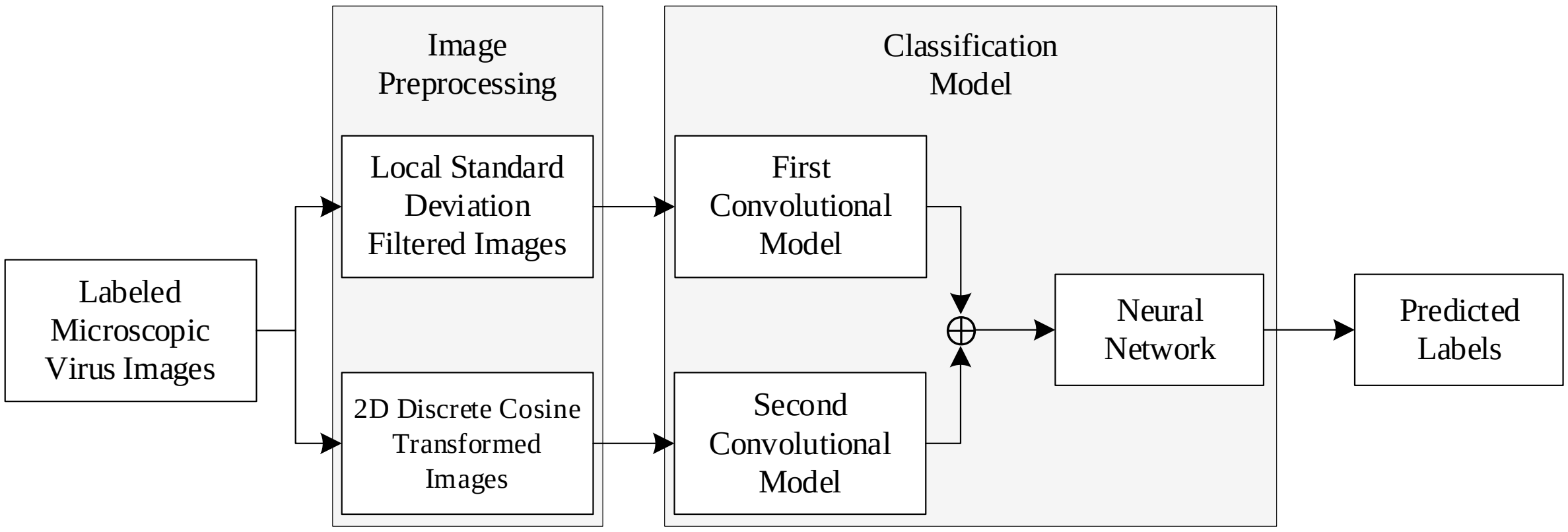

Fig. 1 Methodology of the proposed virus classification scheme.

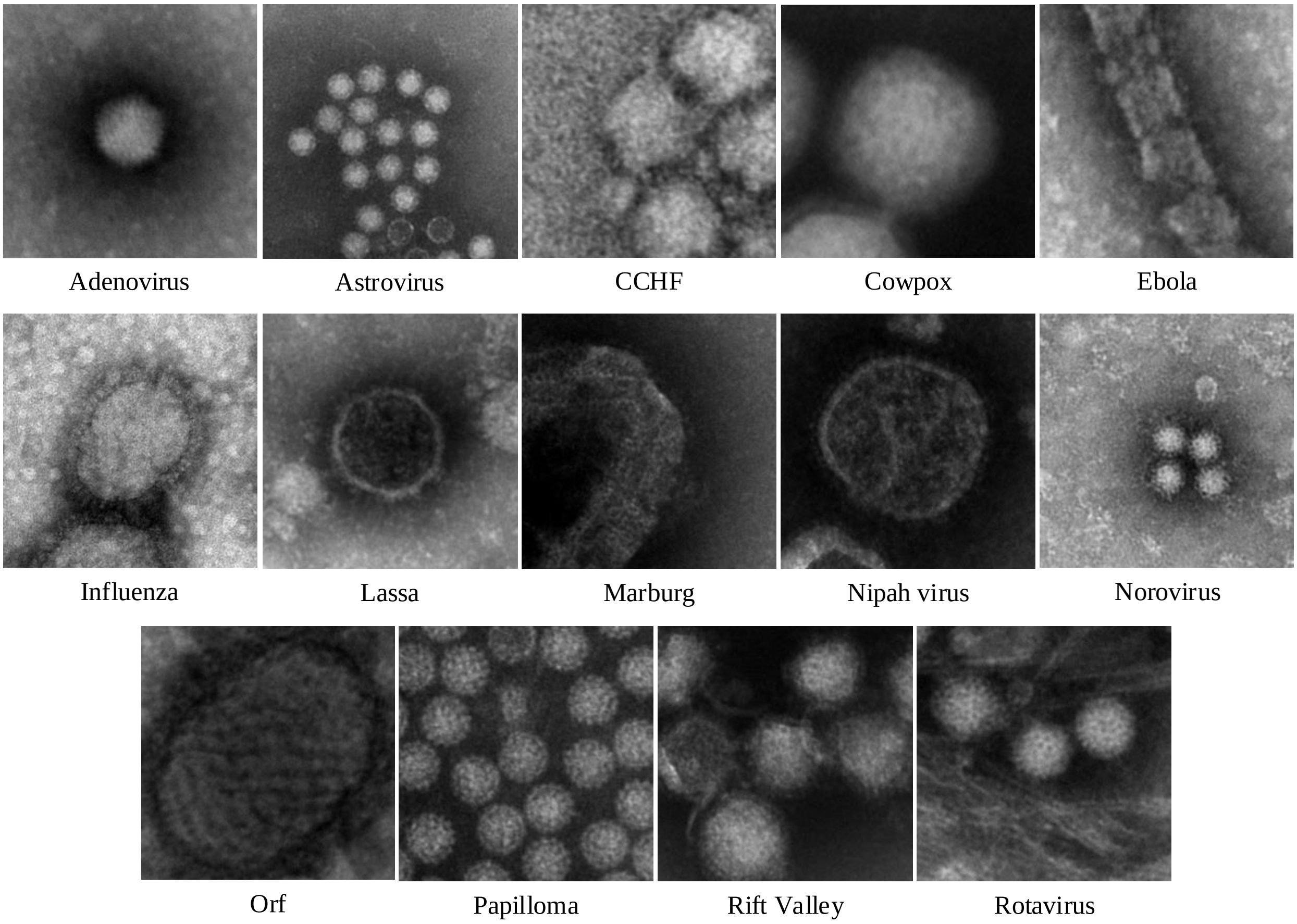

Fig. 2 Sample electron microscopy images of the 14 different kinds of viruses present in the TEM virus dataset [34].

**Table 1** Descriptions of the viruses contained by the TEM virus dataset and used in this study.

| Virus Class | Description and Characteristics | Samples | Class ID |
|---|---|---|---|
| Adenovirus | • a group of medium-sized and non-enveloped viruses capable of affecting people of all ages.<br>• causes many illnesses, including cold-like symptoms, fever, diarrhea, bronchitis, pneumonia, and conjunctivitis. | 736 | *Ad* |
| Astrovirus | • a group of (five or six-pointed) star-shaped viruses that causes diarrhea in humans (primarily infants and young children) and other mammalian species.<br>• other symptoms include nausea, vomiting, stomachache, fever, and loss of appetite. | 736 | *As* |
| CCHF | • causes viral hemorrhagic fever in humans.<br>• usually transmits through ticks from domestic animals.<br>• the mortality rate among victims is relatively high (around 30%).<br>• there is no widely available vaccine for humans or animals. | 736 | *CC* |
| Cowpox | • causes skin infections in numerous species, including humans and domestic animals.<br>• transferable between species through direct contact.<br>• the primary symptom of infection is localized, pustular lesions on the skin. | 736 | *Cp* |
| Ebola | • causes serious viral hemorrhagic fever.<br>• has a high fatality rate (around 50%) and causes occasional outbreaks in Sub-Saharan Africa.<br>• typically transmits to humans and other primates from wild animals. | 736 | *Eb* |
| Influenza | • causes infection in the respiratory system, primarily inside the nose, throat, and lungs.<br>• typical symptoms include runny nose, headache, coughing, sore throat, and fatigue.<br>• children under age five and older adults over age 65 are at high risk of complications. | 736 | *If* |
| Lassa | • causes severe viral hemorrhagic illness that usually lasts from 2 to 21 days.<br>• usually transmits through foods or household items contaminated with rodent urine or feces.<br>• endemic in the West African countries and has a fatality rate of 15% in severe cases. | 736 | *Ls* |
| Marburg | • causes highly virulent hemorrhagic fevers in humans and other primates.<br>• African fruit bat is the reservoir host of this virus, and it usually transmits to humans during visits to mines or caves with bats.<br>• symptoms are similar to Ebola; however, the fatality ratio can be as high as 88%. | 736 | *Mb* |
| Nipah Virus | • causes many complications, including asymptomatic infection, acute respiratory infection, and fatal encephalitis.<br>• can transmit to humans both from animals, such as pigs or bats, and edible items.<br>• estimated mortality rate among victims is between 40% and 75%.<br>• no preventive vaccine or cure is currently available for humans or animals. | 736 | *Np* |
| Norovirus | • causes gastroenteritis (severe inflammation in the stomach and intestine) to the victims.<br>• common symptoms include diarrhea, vomiting, stomach pain, cramps, and low-grade fever.<br>• usually not fatal for adults, but may cause severe dehydration in children and older adults. | 736 | *Nr* |
| Orf | • primarily causes a type of skin infection (usually confined to the epidermis) on the fingers, hands, or forearms.<br>• commonly transmits to humans from sheep and goats through direct contact.<br>• infections may last for multiple weeks and can be very painful, but they usually resolve on their own without leaving scars. | 736 | *Or* |
| Papilloma | • the most common sexually transmitted infection (STI) that causes skin or mucous membrane growths (warts) in the mouth, throat, or genitals.<br>• most transmissions occur through skin-to-skin contact.<br>• although most variants are non-fatal, some can lead to cancers of the anus, cervix, and throat. | 736 | *Pl* |
| Rift Valley | • causes viral fevers in livestock and humans.<br>• transmissions occur through human contact with tissues or body fluids (of infected animals) and bites from mosquitoes carrying the virus. | 736 | *RV* |
| | • usually harmful for animals, not humans; however, in severe cases, victims may experience fever, dizziness, hemorrhage, encephalitis, and eye diseases. | | |
| Rotavirus | • causes highly contagious diarrhea in children and infants, which may lead to dehydration.<br>• symptoms include vomiting, fever, and watery diarrhea that can last from three to eight days.<br>• usually very common and non-fatal, and can be treated at home with extra fluids. | 736 | *Rt* |

To emphasis the benefit of using this filter, we have presented two raw TEM images of the dataset in Fig. 4(a) and Fig. 4(d) alongside the outcomes applying local standard deviation filtering on them in Fig. 4(b) and Fig. 4(e), respectively. As seen from the figures, the edges are more apparent in the latter images, and they look more prominent than their raw counterparts [37]. As convolution highlights boundaries and edges of the shapes or objects present in an image, the images' edge-enhanced versions may carry more significant properties characteristic to their corresponding classes, making them easier to categorize.

## 3.3 Image preparation for the Second Convolutional Model

We performed a domain transformation operation on the raw TEM images and used the output images as the input of the Second Convolutional Model. We used 2D

**Input Image**

| | | | |
|---|---|---|---|
| 40 | 44 | 33 | 34 |
| 43 | 51 | 42 | 44 |
| 45 | 51 | 50 | 50 |
| 42 | 47 | 52 | 53 |

**(a)**

**Symmetric Padded Matrix**

| | | | | | |
|---|---|---|---|---|---|
| 40 | 40 | 44 | 33 | 34 | 34 |
| 40 | 40 | 44 | 33 | 34 | 34 |
| 43 | 43 | 51 | 42 | 44 | 44 |
| 45 | 45 | 51 | 50 | 50 | 50 |
| 42 | 42 | 47 | 52 | 53 | 53 |
| 42 | 42 | 47 | 52 | 53 | 53 |

**(b)**

**3 × 3 Mean values**

| | | | |
|---|---|---|---|
| 42.78 | 41.11 | 39.89 | 36.89 |
| 44.67 | 44.33 | 44.33 | 42.33 |
| 45.44 | 47 | 48.89 | 48.67 |
| 44.78 | 47.56 | 50.56 | 51.78 |

**(c)**

**Local Standard Deviation Filtered Image**

| | | | |
|---|---|---|---|
| 3.56 | 5.62 | 6.55 | 4.88 |
| 4.03 | 5.87 | 7.02 | 7.14 |
| 3.54 | 4.12 | 3.76 | 4.21 |
| 3.15 | 3.97 | 2.30 | 1.39 |

**(d)**

Fig. 3 (a) A sample (partial) TEM image, (b) the image after symmetric padding, (c) mean intensities calculated from 3 × 3 windows, and (d) the outcome of local standard deviation filtering.

DCT for this purpose. Traditional (one-dimensional) DCT converts a signal or data stream (of finite length) to a set of cosine waves oscillating at various frequencies, which (collectively) can express the original signal. Nasir Ahmed first conceived it in 1972 while searching for a better image data compression technique [39,40]. DCT is closely related to Discrete Fourier Transform (DFT), which breaks down a signal into sine and cosine waves. However, DCT has better energy compaction properties than DFT. If we consider a periodic signal, $x(n)$, with finite length $N$, its DCT is defined as [41].

$$x(k) = \alpha(k) \sum_{n=0}^{N-1} x(n) \times \cos\frac{k\pi(2n+1)}{2N}, \qquad (3)$$

$$\alpha(k) = \begin{cases} \frac{1}{\sqrt{N}}, & k = 0 \\ \sqrt{\frac{2}{N}}, & 1 \leq k \leq N-1 \end{cases}$$

For images, which are necessarily 2D matrices, we need 2D DCT. If we consider an image $\boldsymbol{A}$ (with $M \times N$ pixels), its 2D DCT is defined as [42]:

$$\boldsymbol{\mathcal{A}}(k,l) = \alpha(k)\alpha(l) \sum_{m=0}^{M-1} \sum_{n=0}^{N-1} \boldsymbol{A}(m,n) \times \cos\frac{k\pi(2m+1)}{2M} \times \cos\frac{l\pi(2n+1)}{2N} \qquad (4)$$

$$\text{where } \alpha(k) = \begin{cases} \frac{1}{\sqrt{M}}, & k = 0 \\ \sqrt{\frac{2}{M}}, & 1 \leq k \leq M-1 \end{cases} \quad \&$$

$$\alpha(l) = \begin{cases} \frac{1}{\sqrt{N}}, & l = 0 \\ \sqrt{\frac{2}{N}}, & 1 \leq l \leq N-1 \end{cases}$$

2D DCT can be performed in two separate steps, each involving one 1D DCT operation. Apart from image and video compression, DCT is an integral part of various media encoding techniques, communication systems, networking devices, spectral methods, and partial differential equations. Fig. 4(c) and Fig. 4(f) present the 2D DCT outcomes of the microscopic images shown in Fig. 4(a) and Fig. 4(d). Although these images are not very expressive, they contain vital properties of the raw TEM

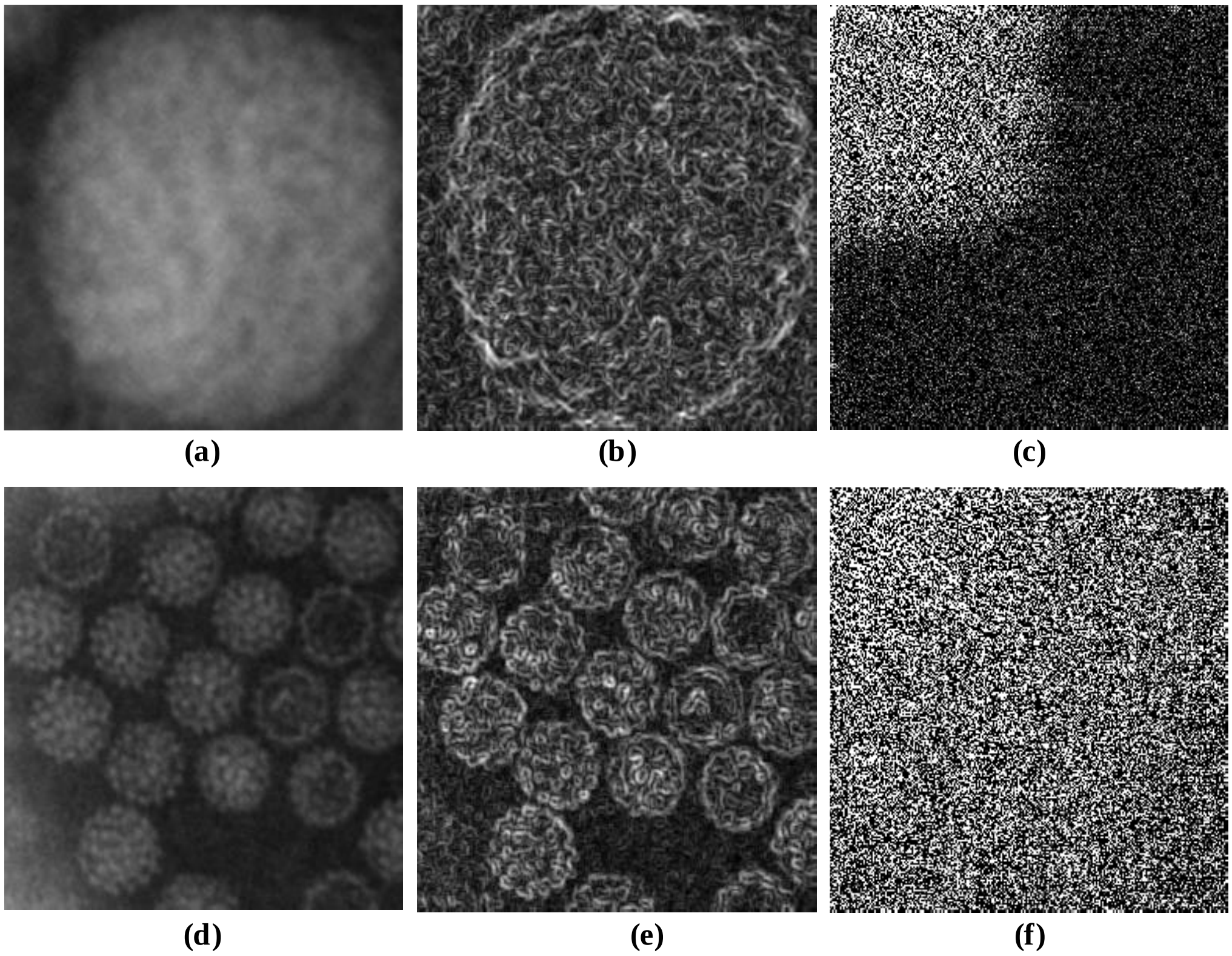

Fig. 4 (a) A sample raw image of the class *Or*, (b) local standard deviation filtered version of (a), (c) 2D DCT output of (a), (d) a sample raw image of the class *Pl*, (e) local standard deviation filtered version of (d), and (f) 2D DCT output of (d).

images and contribute remarkably to the classification model's success, which will be discussed in Section 4.

### 3.4 Architecture of the employed convolutional models and the Neural Network

We used two convolutional models for local feature extraction in the proposed method. After concatenating the sets of features obtained from them, we used a neural network for the final classification. CNNs are feed-forward neural networks, popular and widely used primarily for various image-related tasks, including image classification, reconstruction, compression, and generation. However, it has been modified to work with (data having) lower dimensions (such as signals [43]) and higher dimensions (such as videos [44] and multimodal images [45]) as well. Generally speaking, CNNs are suitable for solving problems that involve the analysis of high-dimensional data with salient spatial structures [46]. CNNs depend on convolution, which mathematically compares an image with a smaller block of image and expresses their similarities (at different regions of the first image) [47]. In this study, our dataset is comprised of grayscale images, meaning they are necessarily 2D matrices with only one (color) channel information. A sample TEM image can be represented as $\boldsymbol{A} \in \mathbb{R}^{m \times n}$, where $m$ and $n$ are the spatial coordinates of $\boldsymbol{A}$. During a convolution, first, $\boldsymbol{A} \in \mathbb{R}^{m \times n}$ goes through an affine transformation. Then, it is passed through an element-wise nonlinear activation. A convolution uses multiple *filters* to extract automatic features from its input data. A 2D filter can be expressed as $\boldsymbol{F}_k \in \mathbb{R}^{s \times s}$, where $s$ is the size of the filter. In this study, we only used $3 \times 3$ square filters for convolution operations; however, larger filter sizes and rectangular-shaped filters can also be used. When applied on $\boldsymbol{A}, \boldsymbol{F}_k$ performs multiple convolutions on the image to obtain a single feature map $\boldsymbol{O}^k$ [46]. The generation of $\boldsymbol{O}^k$ can be expressed as:

$$O_{ij}^{k} = \langle [\boldsymbol{A}]_{ij}, \boldsymbol{F}_k \rangle = \sum_{i'=1}^{s} \sum_{i'=1}^{s} [\boldsymbol{A}]_{i+i'-1,\ j+j'-1} \times [\boldsymbol{F}_k]_{i',j'} \qquad (5)$$

Here $[\boldsymbol{A}]_{ij}$ is a small segment of the image $\boldsymbol{A}$ at location $(i, j)$. $\boldsymbol{F}_k$ travels through the image (from left to right and top to bottom) like a sliding window at a predefined interval in both directions (known as strides, usually set to 1 pixel in each direction) and collects local features from the image. The output of this operation ($\boldsymbol{O}^k$) is a matrix with the size $(m - s + 1) \times (n - s + 1)$ where $[\boldsymbol{O}]_{ijk} = [\boldsymbol{O}^k]_{ij}$. Multiple filters are applied to a single image to get numerous feature vectors like this.

The output of a convolutional layer is passed through a nonlinear activation function. In this study, we used three different activation functions — Sigmoid, Rectified Linear Unit (ReLU), and Softmax. Sigmoid takes a real number ($p$) as its input, and provides a number within [0,1] by computing:

$$f_{Sigmoid}(p) = \frac{1}{1 + e^{-p}} \qquad (6)$$

However, Sigmoid should not be used rapidly, as it often makes the model training process very slow and may kill gradients giving rise to a problem known as "vanishing gradients" [48].

ReLU is probably the most frequently used activation function of recent times because of its simplicity and quick computational ability. A traditional ReLU outputs 0 for a negative input. The input remains unchanged if it is a positive one. Its operation can simply be expressed as:

$$f_{ReLU}(p) = \begin{cases} 0, & p < 0 \\ p, & p \geq 0 \end{cases} \qquad (7)$$

It is best to use ReLU functions only within the hidden layers. Since it ignores the negative inputs completely, it may kill some gradients and often result in dead neurons. There are some modified variants of ReLU that address these problems and reduce their effects.

Softmax is usually used at the last layer of a multi-class classification problem, such as ours. It is also known as multi-class logistic regression since it uses the principles of logistic regression. Mathematically it is defined as:

$$f_{Softmax}(p)_i = \frac{e^{p_i}}{\sum_{j=1}^{J} e^{p_j}} \qquad (8)$$

where $i = 1, 2, 3, \ldots, J$. Softmax is technically a combination of numerous Sigmoid functions, and it can only be used when the dataset classes are mutually exclusive [49].

Apart from the convolution layers, *pooling layers* are also an integral part of CNNs [50]. These layers are used to aggregate all the information of a (defined) window into a single numerical value. Pooling is necessarily a down-sampling operation that reduces the number of available features for the next layers to ease the computational complexity of the model [46]. Although there are several types of pooling layers, max-pooling is the one primarily used. From all the values of a given window (and a given size), max-pooling simply picks the largest value of the window and discards all the others. If a $3 \times 3$ max-pooling window is applied on an image, $\boldsymbol{A} \in \mathbb{R}^{m \times n}$, it will result in a matrix with the size $floor(m/3) \times floor(n/3)$. Although pooling layers are a crucial part of any deep learning model, it is very easy to lose vital information at these layers. Therefore, they should not be used successively or with a large window size.

The convolution and pooling layers necessarily work as feature extractors that draw out classifiable information from the given samples. For classification, CNN uses a feed-forward neural network known as MultiLayer Perceptron (MLP). An MLP is a collection of dense layers, each of which takes its input from its immediately previous layer. The outcome of a dense layer can be defined as:

$$\mathcal{d} = f(W^T \cdot \mathcal{i} + \mathcal{b}) \qquad (9)$$

where $f$ is an activations function, $W \in \mathbb{R}^{m \times n}$ is a weight matrix, $\mathcal{i}$ represents the input, and $\mathcal{b}$ is the bias [51]. The last layer of the MLP outputs a matrix, where each row contains the probabilities of a particular sample to belong at various classes. The sample belongs to the class with the highest probability (as per the model's decision).

As discussed earlier, our method includes two convolutional models. Fig. 5 outlines the architectures of these two models as well as the neural network used for the final classification. As seen from the figure, the First Convolutional Model has three 2D convolution layers, two of which use the Sigmoid activation function, and the remaining one uses the ReLU activation function to process their input data. The Second Convolutional Model, on the other hand, has four 2D convolution layers, and all of them use ReLU activation. In the figure, the arguments of the 2D Convolutions are presented as "number_of _filters, (kernel_size), activation_function." Both models have three max-pooling layers designed to periodically reduce the inputs' size, one batch normalization, and a few dropout layers — which helps to reduce overfitting on the training data. In the figure, the argument of 2D Max-pooling is pool_size; the stride is the same as the pool_size. At the end of each model, a flatten layer brings all the derived features into a vector. At the beginning of the Neural Network block, these two feature sets are concatenated to unify the outputs of both models. The MLP used has five dense layers and a dropout layer to process the mixed features and generate the model's decision. Since the presented architecture processes multiple inputs separately and requires a non-sequential flow of information, it is known as a function deep learning model. The arrangement of the layers and the value of their associated arguments have been devised and finalized upon extensive model hyper tuning specifically for this classification task.

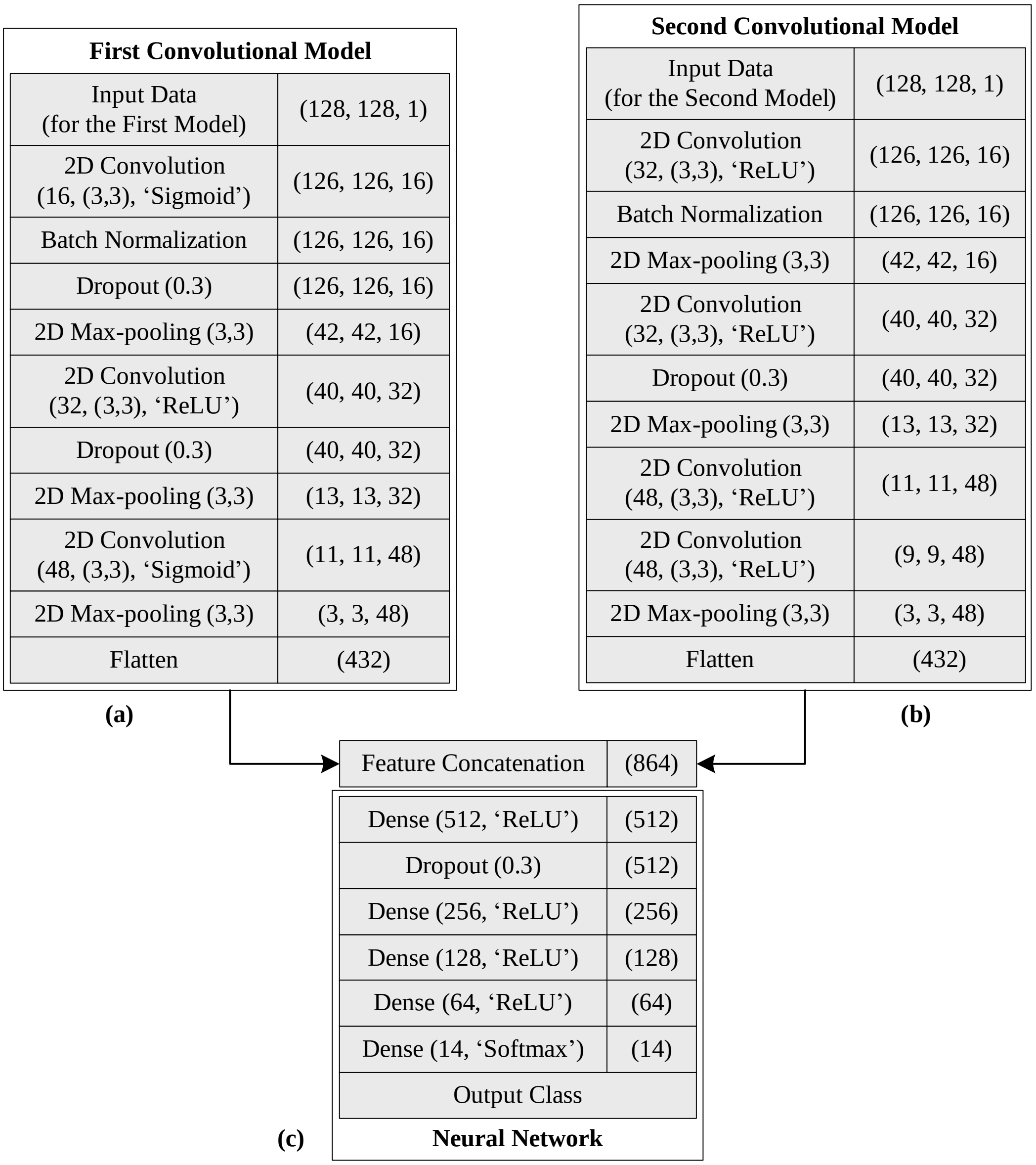

**Fig. 5** (a) Architecture of the (a) First Convolutional Model, (b) Second Convolutional Model, and (c) Neural Network.

## 4 Results and discussion

In this section, we present and delineate the obtained results of the described deep learning model for virus classification. According to the workflow of solving a supervised learning problem, first, the available samples need to be separated into two (in some cases, three) subsets — training and testing. Samples of the training subset are used to teach the model the underlying properties of the images so that it learns to group the samples of the same class and separate among the samples of differing classes. Knowledge acquired by the model at this stage is checked at the latter stage, where the model is given the task of categorizing the testing samples. Finally, the model is judged by comparing its verdict on the samples' labels with their corresponding ground truths (actual labels). In this study, we randomly separated 75% of the samples for the training subset and kept the remaining 25% for testing. As seen from Table 1, the dataset is perfectly balanced. Thus, the model was not biased to any particular class (usually the class with the largest sample size in the case of imbalanced datasets) while making decisions. However, to mitigate the bias even more, we ensured that 75% samples of each class (552 samples) were sent to the training subset. The deep learning model was trained for 100 epochs on an Intel Core i9-9900K Coffee Lake 8-core machine with 32 gigabytes of RAM and 6 gigabytes of GPU. Traditionally the performance of a model is evaluated based on its ability to score the new or unseen (testing) samples correctly. However, we have presented our model's performance on both the training and testing subsets for transparency.

Fig. 6(a) shows the classification accuracy at each epoch of the model's training. Accuracy is the first and foremost evaluation metric used to justify a classifier. It simply expresses the ratio of correctly identified samples and the number of total samples of a classification. As seen from the figure, the highest accuracy on the testing subset was 97.44%, and it was achieved on the 96$^{th}$ epoch. The testing accuracy of the last epoch was 96.43%, which is close to the peak score. The testing accuracies of the 64 epochs were higher than 94%, which indicates excellent classification outcomes during the majority of the epochs. Moreover, the training and testing accuracy curves follow a very similar pattern; the gap between them is reasonably low, which implies a minimal degree of overfitting (on the training samples). The performance was a bit turbulent during the first 20 epochs, but it became pretty stable afterward. At the 96$^{th}$ epoch, the model has a Quadratic Weighted Kappa (QWK) score of 0.9719, which also points out the reliability of the achieved categorical measurements.

Accuracy is just one measurement of the classification outcome, and it can often be misleading, especially while working with an imbalanced dataset. To appraise the results correctly, we need to take a look at some other parameters as well. Precision and recall are two such parameters that answer two vital questions — among the samples identified as a particular class (by the model), how many actually belong to that class, and among all the samples of a given class, how many the model classified correctly. These parameters are calculated individually for each class and then averaged out to get the overall score. Fig. 6(b) and Fig. 6(c) provide the precision and recall scores of the 100 epochs. As seen from these figures, the precision scores were significantly higher during the first few epochs than the corresponding recall scores, hinting below standard classifications. However, the situation improved rather quickly, and from the 10$^{th}$ epoch and onward, these curves followed their corresponding accuracy curves closely. $F_1$-score, another performance evaluation matric that is considered a better descriptor of clarification outcomes than the accuracy score by many, is a weighted average of the previous two metrics. The biggest advantage of explaining a model's performance by its $F_1$-score is that the score does not get affected by class imbalance in the training and testing subsets, although that is not the case for our dataset. Fig. 6(d) presents the $F_1$-score of all the epochs on both subsets. The figure shows the $F_1$-score curves are almost identical to the accuracy curves shown in Fig 6(a), implying that the presented results are accurate and highly reliable.

We have presented the loss curves of the model in Fig. 7(a) to show how the loss of the model decreased gradually with further training. Classification loss expresses how well the classifier has modeled the provided data. It is measured using a specific loss function. A high degree of false predictions causes the loss value to increase. Analysis of the loss values is important because the model itself relies on them (instead of the accuracy or $F_1$-score) to reach the optimal solution with the help of an optimization function. As seen from the figure, the loss values of both subsets decreased gradually with more training. Although the descends of the curves are not smooth, the model reached its optimal state eventually. The curves presented in Fig. 7(b) express the Kullback-Leibler Divergence (KLD) values of the classification. KLD is an asymmetric measure of the distance between two probability distributions, and it is closely related to information divergence and relative entropy [52]. At the 96$^{th}$ epoch, when the model registered its peak performance, the loss and KLD values of classification on the testing subset were 0.0808 and 0.0819, respectively, which are the lowest of the corresponding curves.

In Fig. 8(a) and Fig. 8(b), we have presented the confusion matrices of the best (96$^{th}$ epoch) and the last classifications' outcomes on the testing subset. These matrices can unfold valuable insights into the model's success and failure to identify the samples of each class of the dataset. Confusion matrices are especially useful to determine the particular classes the method is struggling to recognize. As discussed above, we kept 184 samples of each class in the testing subset (2576 in total). Fig. 8(a) shows that the model misclassified only 66 samples among them at the best epoch. Among the 14 classes, *Or* and *RV* have the highest correct classification rate and the highest misclassification rate, respectively. Class-wise precision, recall, and $F_1$-scores have been calculated from this confusion matrix and presented in Table 2. It is easier to grasp the individual class performances from (the table's) numerical values. It shows that apart from *Or*, the model had near-perfect classification rates in *As, Cp*, *Nr*, *Pl,* and *Rt*. On the other hand, the model's least successful performance laid in *Eb*, *Mb*, and *RV*. However, 10 of the 14 classes had $F_1$-scores above 97%, which implies that the model was proficient in categorizing samples of most of the classes. Table 2 also contains the Area Under Curve (AUC) scores (of the classes) measured for drawing their corresponding Receiver Operating Characteristic (ROC) curves. AUC expresses the ratio of the area occupied by the associated class's ROC curve and the total area of the ROC graph's 2D space. As all the classes had AUC scores above 99%, we can safely conclude that their ROC curves will be near-perfect as well. The confusion matrix of the last epoch, presented in Fig. 8(b), almost mimics its counterpart, only 26 more samples were misclassified than the previous instance. Nevertheless, the last epoch's performance is quite similar to that of the best one.

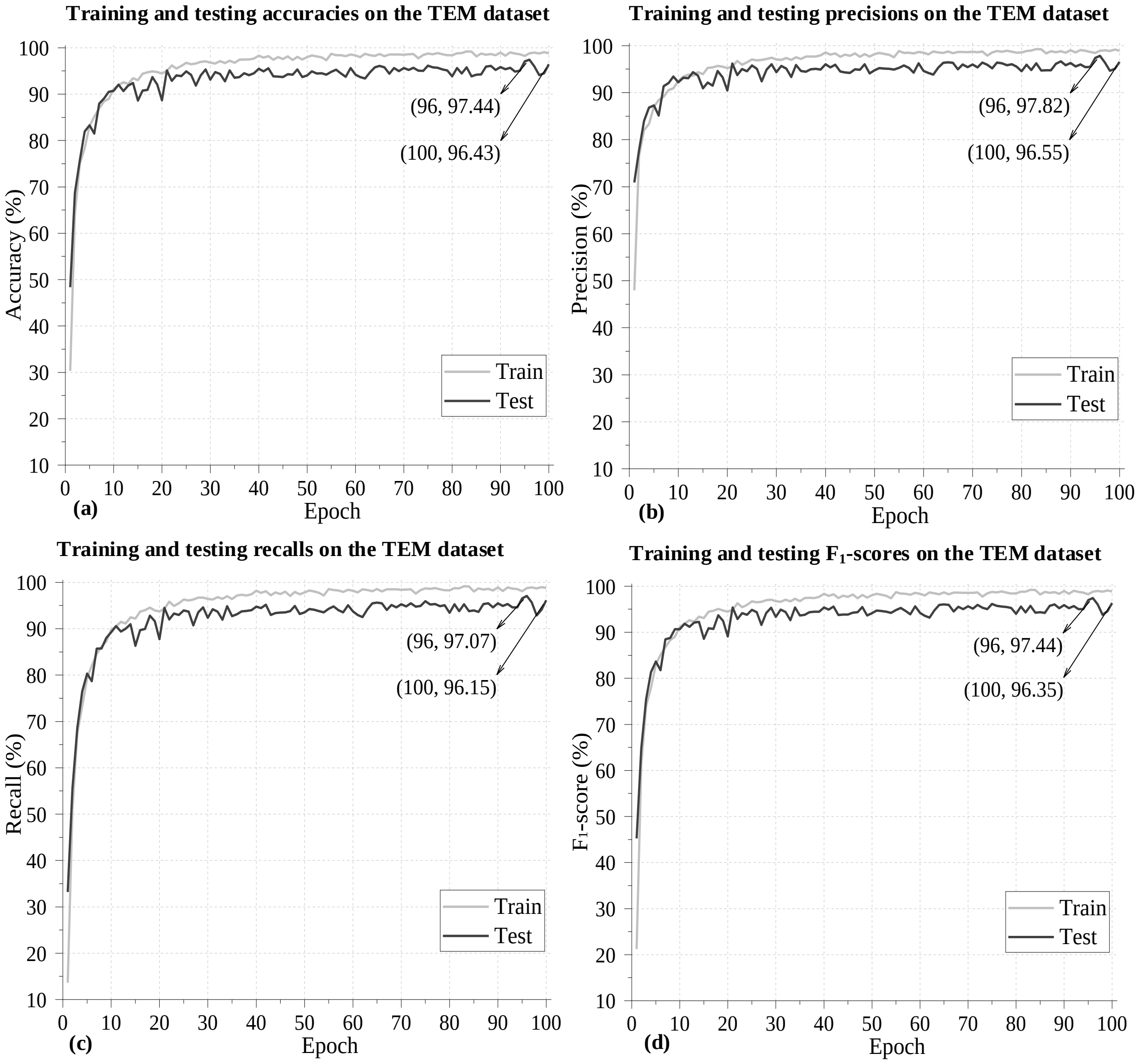

**Fig. 6** The performance of the proposed model on the TEM Virus dataset in terms of the obtained classification (a) accuracy, (b) precision, (c) recall, and (d) F1-score on both the training and testing subsets.

In the previous paragraphs, we have described the results of the proposed model. However, as seen from Fig. 1, it has two visibly separate pipelines where two sets of images were created and processed with two different convolutional models. In the interest of fairness and clarity, we have presented the same classification results with only one pipeline (one image pre-processing and one convolutional model for feature extraction) at a time in Table 3. The first experiment involves the imaging technique outlined in Section 3.2, the First Convolutional Model portrayed in Fig. 5(a), and the Neural Network from Fig. 5(c). Similarly, the second experiment involves 2D DCT described in Section 3.3, the Second Convolutional Model portrayed in Fig. 5(b), and the same Neural Network. As seen from the table, the latter one provided notably better results than the first setup. However, the described model attained an even better performance by combining the outputs of the two pipelines. It registered a 5.24% increase in accuracy and 5.32% increase in the average $F_1$-score on the second experiment yielding the benefits of incorporating two sets of features extracted from the raw images.

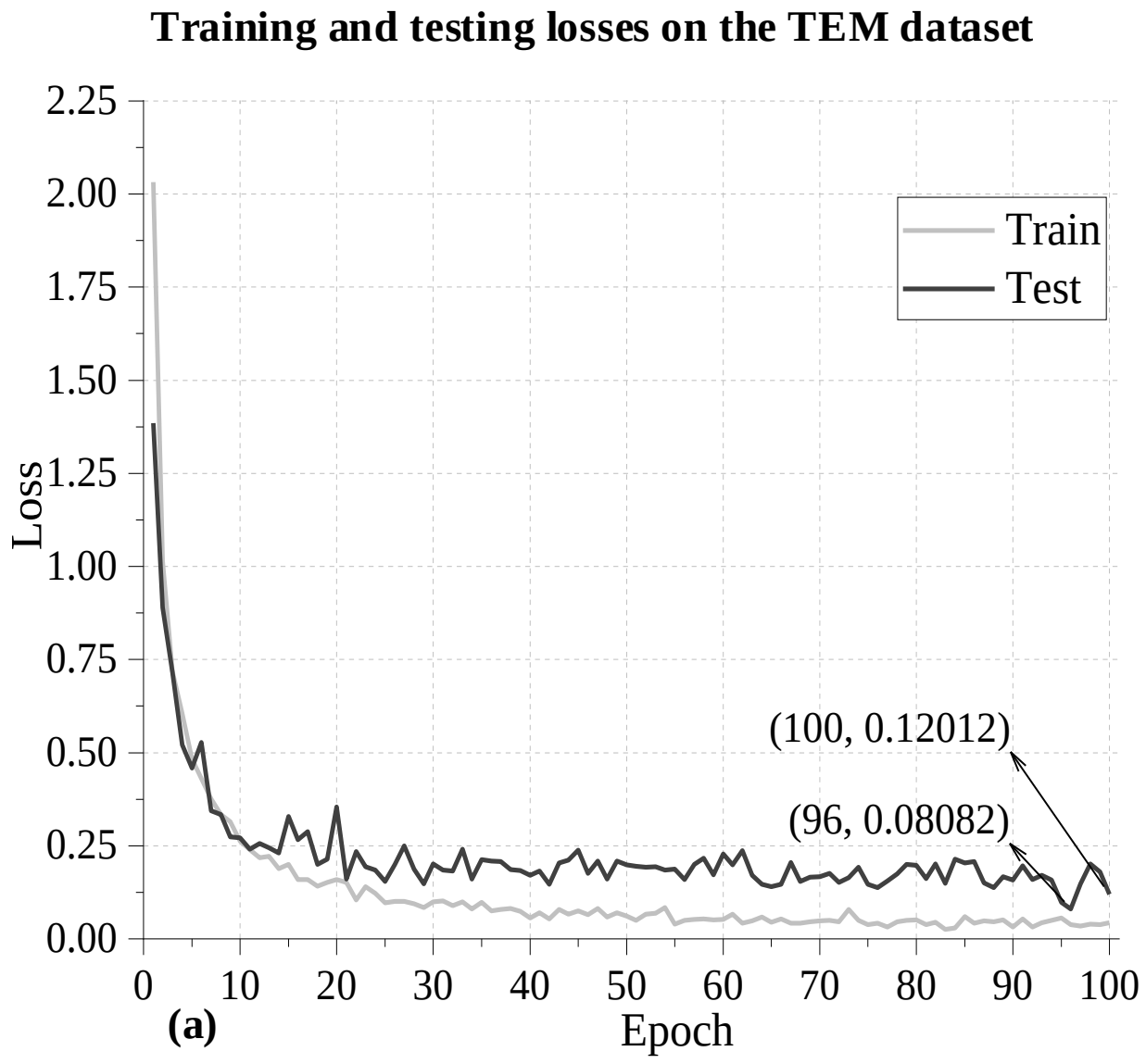

**Fig. 7** The (a) loss and (b) KLD curves of the classifications on the training and testing subsets.

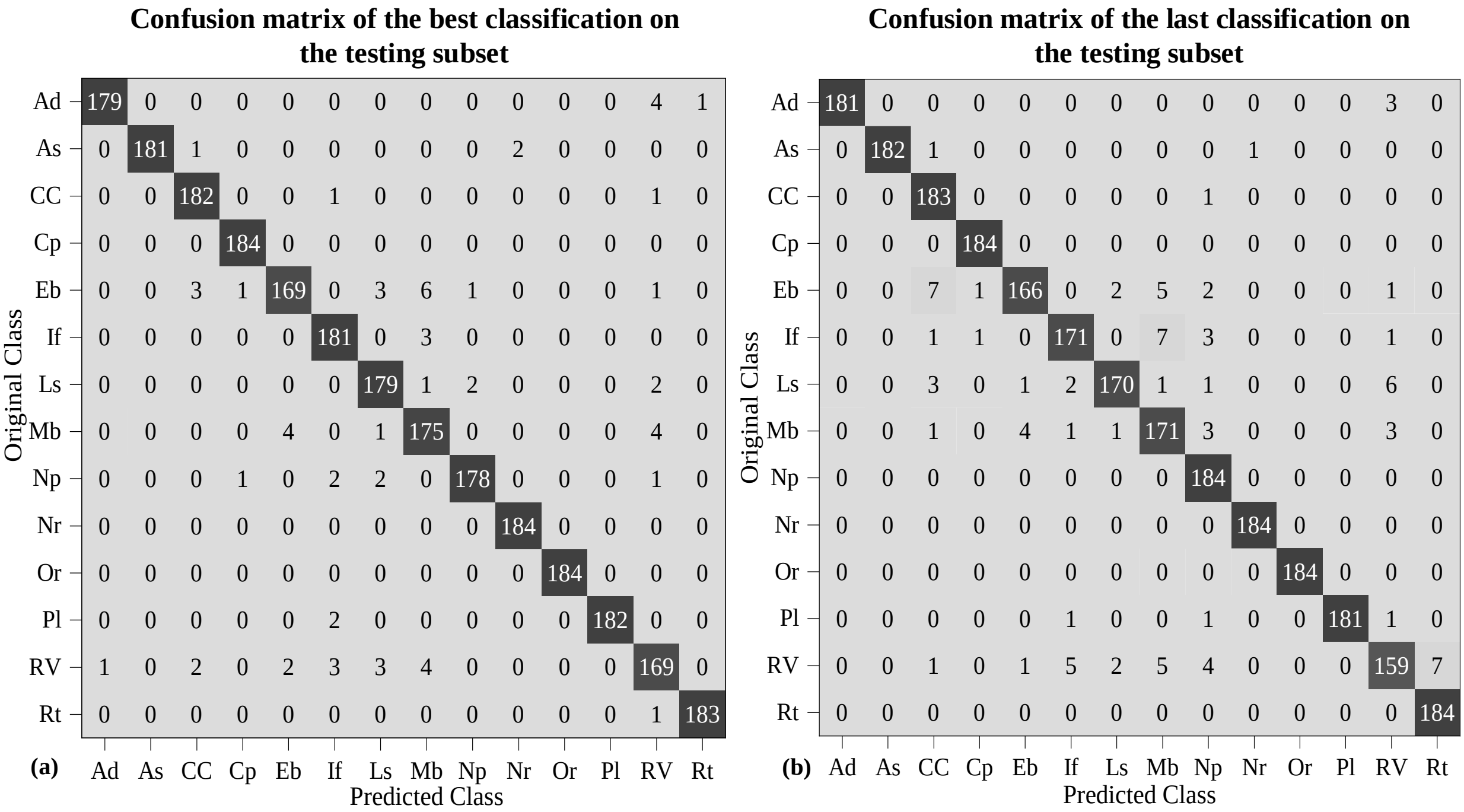

**Fig. 8** The confusion matrix of the (a) best (96th) and (b) last classification on the testing subset.

To put the performance acquired by our method into perspective, we have compared it with other similar studies (discussed in Section 2) in Table 4. It is essential to mention that we used a very recent dataset (published in 2021) for this study to train and validate our model. In contrast, most of the mentioned studies were done on the Virus Texture Dataset. In that sense, our results are not directly comparable with the results of these studies. However, we have made the comparison bearing in mind that the end goal is the same — to differentiate among numerous virus species as accurately and reliably as possible. The table shows that the highest accuracy reported on the Virus Texture Dataset is 95.40%, achieved by [31]. Our study has outperformed it by over 2% in terms of the accuracy score. However, further quantitative comparisons are impossible since most of these studies have not reported any other scores (such as precision, recall, or $F_1$-score). Furthermore, some studies (especially the ones that used non-deep learning methods for classification) have presented the mean accuracy score of multiple classifications. In contrast, others have claimed the best accuracy they achieved. This makes a comparative discussion even more challenging. However, the results presented in [35] and [36] are directly comparable to ours as both are based on the same dataset. In [35], the authors reported a 93.10% classification accuracy, 92.60% precision, 92.10% recall, and 92.10% $F_1$-score on the same image samples. As seen in Table 3, our results in comparison are 4.34%, 5.22%, 4.97%, and 5.34% higher in the respective metrics. The CNN-based model presented in [36] achieved a 96.10% classification accuracy, which is less than that of the proposed method.

Therefore, based on all the presented results, it can be concluded that the method described here is highly effective and reliable in TEM image-based virus classification. The TEM virus dataset has quite a few advantages over the benchmark Virus Texture Dataset, including increased samples (in each class) and higher image resolutions. We hope more studies will use this dataset and report their acquired results in the coming years.

**Table 2** Class-wise classification performance.

| Class ID | Precision (%) | Recall (%) | $F_1$-score (%) | AUC (%) |
|---|---|---|---|---|
| *Ad* | 99.44 | 97.28 | 98.35 | 99.99 |
| *As* | **100** | 98.37 | 99.18 | **100** |
| *CC* | 96.81 | 98.91 | 97.85 | 99.99 |
| *Cp* | 98.92 | **100** | 99.46 | **100** |
| *Eb* | 96.57 | 91.85 | 94.15 | 99.84 |
| *If* | 95.77 | 98.37 | 97.05 | 99.98 |
| *Ls* | 95.21 | 97.28 | 96.24 | 99.95 |
| *Mb* | 92.59 | 95.11 | 93.83 | 99.88 |
| *Np* | 98.34 | 96.74 | 97.53 | 99.97 |
| *Nr* | 98.92 | **100** | 99.46 | **100** |
| *Or* | **100** | **100** | **100** | **100** |
| *Pl* | **100** | 98.91 | 99.45 | **100** |
| *RV* | 92.35 | 91.85 | 92.10 | 99.64 |
| *Rt* | 99.46 | 99.46 | 99.46 | **100** |
| Average | 97.82 | 97.07 | 97.44 | 99.95 |

## 5 Conclusions

In this article, we proposed and inscribed a novel heterogeneous virus classification method that takes information from electron microscopic images of the viruses. The data preparation stage involves two common image processing techniques, namely local standard deviation filtering and two-dimensional DCT, to prepare TEM images for automatic feature extraction. The deep-learning model used for classification consists of two convolutional architectures and a neural network customized for this task. The model's performance has been expressed with multiple evaluation matrices, all of which yield that the method is highly successful in identifying the samples of the majority of the virus classes. However, the model's performance can improve for at least four classes' samples. Overall, the method registered a 97.44% peak classification accuracy and F1-score, and a QWK score of 0.9719, on the samples of the testing subset — which can be labeled as excellent classification outcomes. Comparison with other contemporary studies shows that our method is considerably better at distinguishing among various types of viruses through analyzing their TEM images. We hope that this fast and automatic identification model will be practically implemented, and its decisions will be taken into consideration by the virologists and other medical professionals to reach a more robust verdict while diagnosing a possible patient. In the future, we would like the focus on this model's limitations and work on devising ways to mitigate them. Experimenting with different image filtering and noise-removal methods, updating the convolutional models to extract more characteristic features, and changing the neural network to lessen the degree of overfitting might be some of our possible endeavors.

**Table 3** Comparison of performance using different preprocessing methods and convolutional models.

| Preprocessing | Model | Accuracy (%) | Precision (%) | Recall (%) | $F_1$-score (%) |
|---|---|---|---|---|---|
| Local Standard Deviation Filtering | First Convolutional Model + Neural Network | 76.67 | 78.76 | 75.42 | 77.05 |
| 2D DCT | Second Convolutional Model + Neural Network | 92.20 | 92.35 | 91.89 | 92.12 |
| Both | Proposed | **97.44** | **97.82** | **97.07** | **97.44** |

**Table 4** Comparison of the acquired performance with similar methods.

| Study | Year | Data Source | Classes | Main Classifier | Accuracy (%) |
|---|---|---|---|---|---|
| [20] | 2014 | Virus Texture Dataset [19] and non-virus data | 16 | RF | 89.00 |
| [21] | 2014 | Virus Texture Dataset | 15 | SVM | 85.70* |
| [22] | 2014 | Virus Texture Dataset | 15 | LBP | 67.50* |
| [23] | 2016 | Virus Texture Dataset | 15 | SVM | 86.20* |
| [24] | 2017 | Virus Texture Dataset | 5 | FFNN | 88.00 |
| [25] | 2018 | Virus Texture Dataset | 15 | SVM | 88.00 |
| [26] | 2019 | Virus Texture Dataset | 16 | U-Net | 82.20* |
| [27] | 2019 | Virus Texture Dataset | 15 | SVM | 82.07* |
| [32] | 2020 | Various | 5 | NB | 76.70* |
| [28] | 2020 | Virus Texture Dataset | 15 | SVM | 87.27* |
| [29] | 2020 | Virus Texture Dataset | 15 | DenseNet | 89.00* |
| [30] | 2020 | Virus Texture Dataset | 15 | DenseNet201 | 89.47 |
| [35] | 2021 | TEM virus dataset | 14 | DenseNet201 | 93.10 |
| [36] | 2022 | TEM virus dataset | 14 | CNN | 96.10 |
| Proposed | 2023 | TEM virus dataset | 14 | CNN | **97.44** |

*Mean accuracy score (presented in the corresponding study).

**Acknowledgements** The authors would like to cordially thank Damian Matuszewski, Ida-Maria Sintorn, and their associates for creating the TEM virus dataset and making it available for public use.

**Competing Interests** The authors declare that they have no competing interests.

**Funding** This research received no funding.

**Data Availability** This research used a publicly available dataset. Processed data and codes will be available on request.